%% file: top.tex
\documentclass[10pt,twocolumn,letterpaper]{article}

\usepackage{cvpr}
\usepackage{times}
\usepackage{epsfig}
\usepackage{graphicx}
\usepackage{amsmath}
\usepackage{amssymb}
\usepackage{multirow}

\usepackage[pagebackref=true,breaklinks=true,letterpaper=true,colorlinks,bookmarks=false]{hyperref}

\cvprfinalcopy 

\ifcvprfinal\fi
\begin{document}

\title{PIXOR:  Real-time 3D Object Detection from Point Clouds}

\author{Bin Yang\quad Wenjie Luo\quad Raquel Urtasun\\
Uber Advanced Technologies Group\quad University of Toronto\\
{\tt\small \{byang10, wenjie, urtasun\}@uber.com}}

\maketitle

\input{abs}
\input{intro}
\input{related}
\input{model}
\input{results}
\input{conc}
\input{ack}

{\footnotesize
\bibliographystyle{ieee}
\bibliography{egbib}
}

\end{document}

%% file: abs.tex
\begin{abstract}

We address the problem of real-time 3D object detection from point clouds in the context of autonomous driving. 
Computation speed is critical as detection is a necessary component for safety. 
Existing approaches are, however, expensive in computation due to high dimensionality of point clouds. 
We utilize the 3D data more efficiently by representing the scene from the Bird's Eye View (BEV), and propose PIXOR, a proposal-free, single-stage detector that outputs oriented 3D object estimates decoded from pixel-wise neural network predictions.
The input representation, network architecture, and model optimization are especially designed to balance high accuracy and real-time efficiency.
We validate PIXOR on two datasets: the KITTI BEV object detection benchmark, and a large-scale 3D vehicle detection benchmark. In both datasets we show that the proposed detector surpasses 
other state-of-the-art methods notably in terms of Average Precision (AP), while still runs at $>28$ FPS.

\end{abstract}

%% file: intro.tex
\section{Introduction}

Over the last few years we have seen a plethora of methods that exploit Convolutional Neural Networks to produce accurate 2D object detections, typically from a single image \cite{rcnn, frcn, faster, rfcn, yolo, ssd}.
However, in robotics applications such as autonomous driving we are interested in detecting objects in 3D space. This is fundamental for motion planning in order to plan a safe route.

Recent approaches to 3D object detection exploit different data sources.
Camera based approaches utilize either monocular \cite{mono3d} or stereo images \cite{3doppami}. However, accurate 3D estimation from 2D images is difficult, particularly in long ranges. 
With the popularity of inexpensive RGB-D sensors such as Microsoft Kinect, Intel RealSense and Apple PrimeSense, several approaches that utilize depth information and fuse them with RGB images have been developed \cite{ss, dss}. They have been shown to achieve significant performance gains over monocular methods.  In the context of autonomous driving, high-end sensors like LIDAR (Light Detection And Ranging) are more common because higher accuracy is needed for safety. 
The major difficulty in dealing with LIDAR data is that the sensor produces unstructured data in the form of a point cloud containing typically around $10^5$ 3D points per 360-degree sweep. This poses a large computational challenge for modern detectors. 

Different forms of point cloud representation have been explored in the context of 3D object detection. The main idea is to form a structured representation where standard convolution operation can be applied.  Existing representations are mainly divided into two types: 3D voxel grids and 2D projections.
A 3D voxel grid transforms the point cloud into a regularly spaced 3D grid, where each voxel cell can contain a scalar value (e.g., occupancy) or vector data (e.g., hand-crafted statistics computed from the points within that voxel cell). 
3D convolution is typically applied to extract high-order representation from the voxel grid \cite{vote3deep}.
However, since point clouds are sparse by nature, the voxel grid  is  very sparse and therefore a large proportion of computation is redundant and unnecessary. As a result, typical systems that use this representation \cite{vote3deep, vote3d, velofcn} only run at 1-2 FPS.

An alternative is to project the point cloud onto a plane, which is then discretized into a 2D image based representation where 2D convolutions are applied. During discretization, hand-crafted features (or statistics) are computed as pixel values of the 2D image \cite{mv3d}. 
Commonly used projections are range view (i.e., 360-degree panoramic view) and bird's eye view (i.e., top-down view).
These 2D projection based representations are more compact, but they bring information loss during projection and discretization. For example, range-view projection will have distorted object size and shape. To alleviate the information loss, MV3D \cite{mv3d} proposes to fuse the 2D projections with the camera image to bring additional information. However, the fused model has nearly linear computation cost with respect to the number of input modalities, making real-time application infeasible.

\begin{figure*}[t]
\begin{center}
   \includegraphics[width=1.0\linewidth]{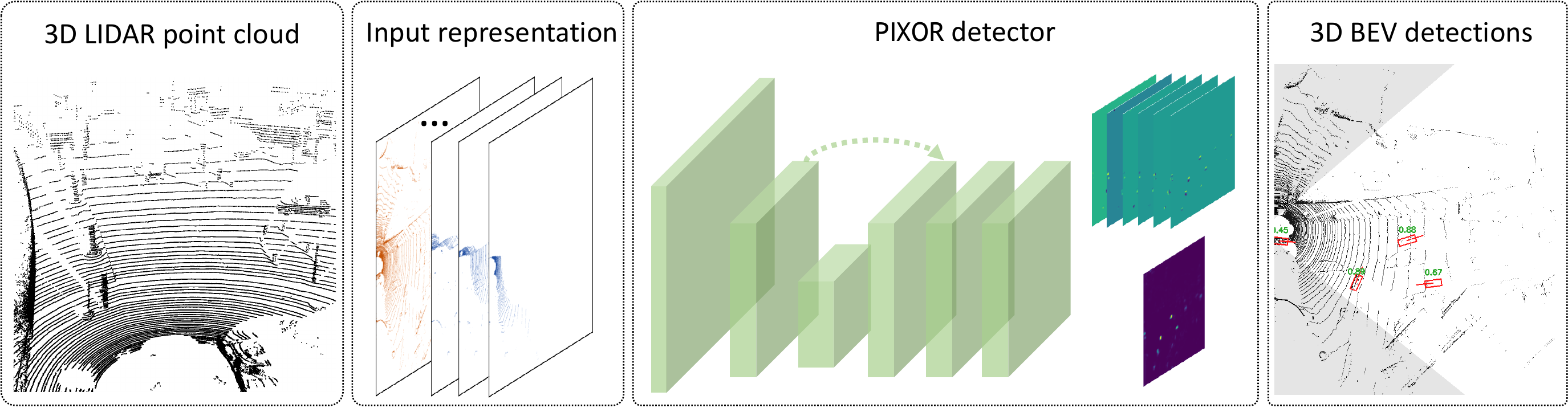}
\end{center}
   \caption{Overview of the proposed 3D object detector from Bird's Eye View (BEV) of LIDAR point cloud.}
\label{fig:trailer}
\end{figure*}

In this paper, we propose an accurate {\it real-time} 3D object detector, which we call {\bf PIXOR} ({\bf OR}iented 3D object detection from {\bf PIX}el-wise neural network predictions), that operates on 3D point clouds.
PIXOR is a single-stage, proposal-free dense object detector that exploits the 2D Bird's Eye View (BEV) representation in an efficient way.
We choose the BEV representation as it is computationally more friendly compared with 3D voxel grids, and also preserves the metric space which allows our model to explore priors about the size and shape of the object categories. 
Our detector outputs accurate oriented bounding boxes in real-world dimensions in bird's eye view. Note that these are 3D estimates as we assume that the objects are on the ground. This is a reasonable assumption in the autonomous driving scenario as vehicles do not fly. 

We demonstrate the effectiveness of our approach in two datasets, the public KITTI benchmark \cite{kitti} and a large-scale 3D vehicle detection dataset TOR4D.
Specifically, PIXOR achieves the highest Average Precision (AP) on KITTI bird's eye view object detection benchmark among all previously published methods, while also runs the fastest among them (over 28 FPS). We also provide in-depth ablation studies on KITTI to investigate how much performance gain each module contributes, and prove the scalability and generalization ability of PIXOR by applying it to the large-scale TOR4D dataset.

%% file: related.tex
\section{Related Work}

We first review recent advances in applying Convolutional Neural Networks to object detection, and then revisit works in two related sub-fields, single-stage object detection and 3D object detection.

\subsection{CNN-based Object Detection}
Convolutional Neural Networks (CNN) have shown outstanding performance in image classification \cite{alexnet}. When applied to object detection, it is natural to utilize them by running inference over cropped regions representing the object candidates. Overfeat \cite{overfeat} slides a CNN on different positions and scales and predicts a bounding box per class at each time. Since the introduction of class-agnostic object proposals \cite{selective, mcg}, proposal based approaches become more popular, with Region-CNN (RCNN) \cite{rcnn} and its faster versions \cite{frcn, rfcn} being the most seminal work. RCNN first extracts the whole-image feature map with an ImageNet \cite{deng2009imagenet} pre-trained CNN and then predicts a confidence score as well as box position per proposal via an RoI-pooling operation on the whole-image feature map \cite{spp}. Faster-RCNN \cite{faster} further proposes to learn to generate region proposals with a CNN and share the feature representation with detection, which leads to further gains in both performance and speed. Proposal based object detectors achieve outstanding performances in many public benchmarks \cite{everingham2010pascal, russakovsky2015imagenet}. However, the typical two-stage pipeline makes it unsuitable for real-time applications.

\subsection{Single-stage Object Detection}
Different from the two-stage detection pipeline that first predicts proposals and then refines them, single-stage detectors directly predict the final detections. YOLO \cite{yolo} and SSD \cite{ssd} are the most representative works with real-time speed. YOLO \cite{yolo} divides the image into sparse grids and makes multi-class and multi-scale predictions per grid cell.  SSD \cite{ssd} additionally uses pre-defined object templates (or \emph{anchors}) to handle the large variance in object size and shape. For single-class object detection, DenseBox \cite{densebox} and EAST \cite{east} show that single-stage detector also works well without using manually designed anchors. They both adopt the fully-convolutional network architecture \cite{long2015fully} to make dense predictions, where each pixel location corresponds to one object candidate. Recently RetinaNet \cite{focal} shows that single-stage detector can outperform two-stage detector if class imbalance problem during training is resolved properly. Our proposed detector follows the idea of single-stage dense object detector, while further extends these ideas to real-time 3D object detection by re-designing the input representation, network architecture, and output parameterization. We also remove the ``pre-defined object anchors'' hyper parameter by re-defining the objective function of object localization, which leads to a simpler detection framework.

\subsection{3D Object Detection from Point Clouds}
Vote3D \cite{vote3d} uses sliding window on sparse volumes in a 3D voxel grid to detect objects. Hand-crafted geometry features are extracted on each volume and fed into an SVM classifier \cite{suykens1999least}. Vote3Deep \cite{vote3deep} also uses the voxel representation of point clouds, but extracts features for each volume with 3D CNN \cite{tran2015learning}. The main issue with voxel representations is efficiency, as the 3D voxel grid usually has high dimensionality. In contrast, VeloFCN \cite{velofcn} projects the 3D point cloud to front-view and gets a 2D depth map. Vehicles are then detected by applying a 2D CNN on the depth map. Recently MV3D \cite{mv3d} also uses the projection representation. It combines CNN features extracted from multiple views (front view, bird's eye view as well as camera view) to do 3D object detection. However, hand-crafted features are computed as the encoding of the rasterized images. Our proposed detector, however, uses the bird's eye view representation alone for real-time 3D object detection in the context of autonomous driving, where we assume that all objects lie on the same ground.

%% file: model.tex
\section{PIXOR Detector}

In this paper we propose an efficient 3D object detector that is able to produce very accurate bounding boxes given LIDAR point clouds. Our bounding box estimates not only contain the location in 3D space, but also the heading angle, since predicting this accurately is very important for autonomous driving. 
We exploit a 2D representation of LIDAR point clouds, as it is more compact and thus amenable to real-time inference compared with 3D voxel grid representation. 
An overview of the proposed 3D object detector is shown in Figure \ref{fig:trailer}. In the following we introduce our input representation, network architecture and discuss how we encode the oriented bounding boxes. We also present details about the learning of and inference with the detector.
 
\subsection{Input Representation}

Standard convolutional neural networks perform discrete convolutions and thus assume that the input lies on a grid. 
3D point clouds are however unstructured, and thus standard convolutions cannot be directly applied. 
One option is to use voxelization to form a 3D voxel grid, where each voxel cell contains certain statistics of the points that lie within that voxel. To extract feature representation from this 3D voxel grid, 3D convolution is often used. However, this can be very expensive in computation as we have to slide the 3D convolution kernel along three dimensions. This is also unnecessary because the LIDAR point cloud is so sparse that most voxel cells are empty. 

Instead, we can represent the scene from the bird's eye view (BEV) alone. By reducing the free degrees from 3 to 2, we don't lose information in point cloud as we can still keep the height information as channels along the third dimension (like the RGB channels of 2D images). However, effectively we get a more compact representation since we can apply 2D convolution to the BEV representation. This dimension reduction is reasonable in the context of autonomous driving as the objects of interest are on the same ground. In addition to computation efficiency, BEV representation also have other advantages. It eases the problem of object detection as objects do not overlap with each other (compared with front-view representation). It also keeps the metric space, and thus the network can exploit priors about the physical dimensions of objects. 

Commonly used features for voxelized LIDAR representation are occupancy, intensity (reflectance), density, and height feature \cite{mv3d}. In PIXOR, for simplicity we only use occupancy and intensity as the features. 
In practice, we first define the 3D physical dimension $L \times W \times H$ of the scene that we are interested in. We then compute occupancy feature maps at a grid resolution of $d_{L} \times d_{W} \times d_{H}$, and compute the intensity feature map at a grid resolution of $d_{L} \times d_{W} \times H$. Note that we add two additional channels to occupancy feature maps to cover out-of-range points. The final representation has the shape of $\frac{L}{d_L} \times \frac{W}{d_W} \times (\frac{H}{d_H} + 3)$.

\begin{figure}[t]
\begin{center}
   \includegraphics[width=1.0\linewidth]{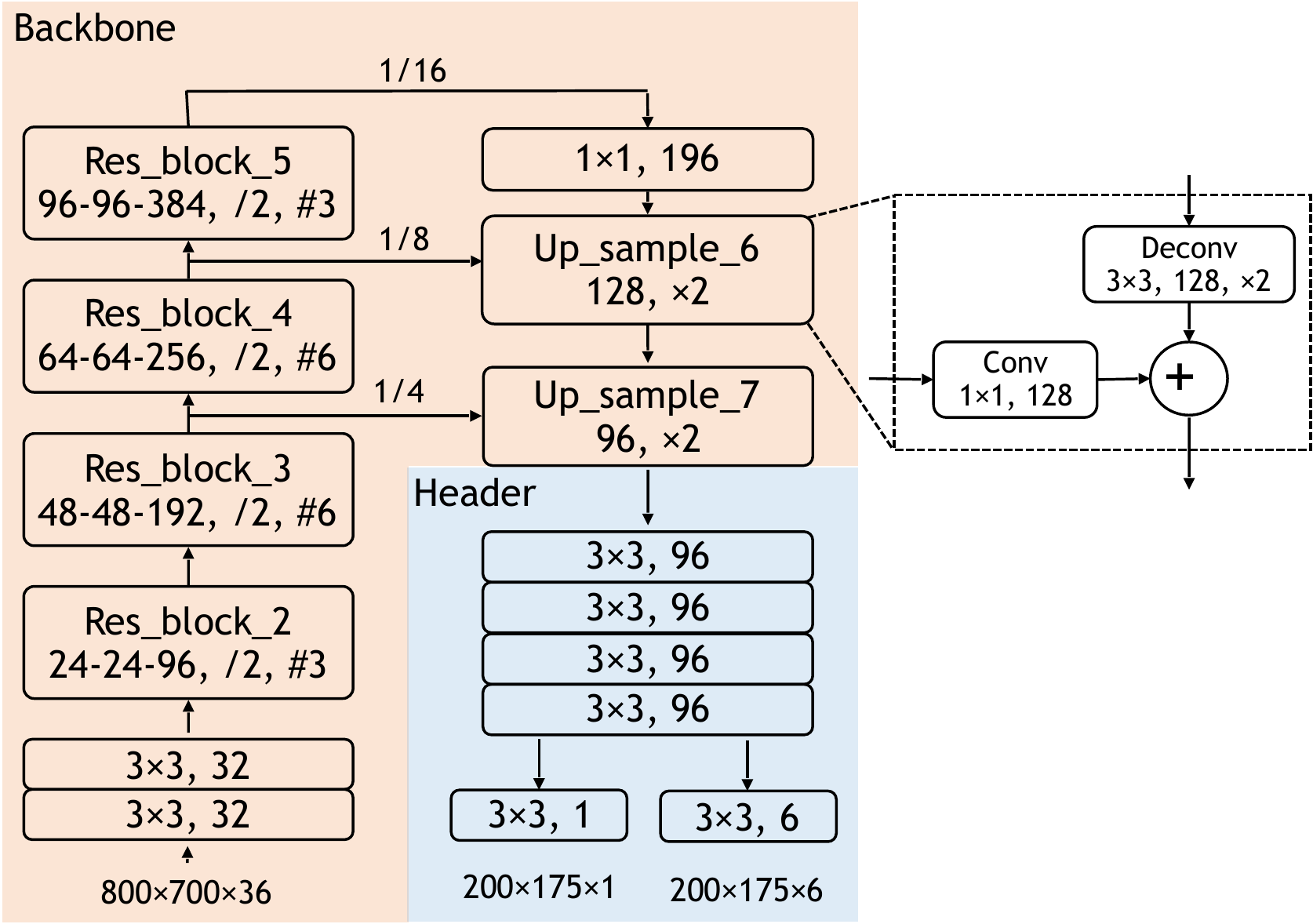}
\end{center}
   \caption{The network architecture of PIXOR.}
\label{fig:network_architect}
\end{figure}

\subsection{Network Architecture}

PIXOR uses a fully-convolutional neural network designed for dense oriented 3D object detection. We do not adopt the commonly used proposal generation branch \cite{frcn, faster, rfcn, mv3d}. Instead, the network outputs pixel-wise predictions at a single stage, with each prediction corresponding to a 3D object estimate. As a result the recall rate of PIXOR is $100\%$ by definition. Thanks to the fully-convolutional architecture, such dense predictions can be computed very efficiently.  In terms of the encoding of the 3D object in the network prediction, we use the direct encoding without resorting to pre-defined object anchors \cite{frcn, faster, rfcn}, which works well in practice.
All these designs make PIXOR extremely simple and generalize well thanks to zero hyper-parameter in network architecture. To be specific, there is no need to design object anchors, nor to tune the number of proposals passed from the first stage to the second stage along with the corresponding Non-Maximum-Suppression threshold.

We show the architecture of PIXOR in Figure \ref{fig:network_architect}. The whole architecture can be divided into two sub-networks: a \emph{backbone network} and a \emph{header network}. The {\it backbone network} is used to extract general representation of the input in the form of convolutional feature maps. It has high representation capacity to learn a robust feature representation. The {\it header network} is used to make task-specific predictions, and in our case it has a single-branch structure with multi-task outputs: a score map representing the object class probability, and the geometry maps encoding the size and shape of the oriented 3D objects.

\subsubsection{Backbone Network}
Convolutional Neural Networks are typically composed of convolutional layers and pooling layers. Convolutional layers are used to extract an over-complete representation of the input feature, while pooling layers are used to down-sample the feature map size to save computation and help create more robust representation. The backbone networks in many image based object detectors usually have a down-sampling factor of $16$ \cite{faster, frcn, rfcn}, and are typically designed to have fewer layers in high-resolution and more layers in low-resolution. It works well for images as objects are typically large in pixel size. However, this will cause a problem in our case as objects can be very small. A typical vehicle has a size of $18 \times 40$ pixels when using a discretization resolution of $0.1m$. After $16\times$ down-sampling, it covers around $3$ pixels only.

One direct solution is to use fewer pooling layers. However, this will decrease the size of the receptive field of each pixel in the final feature map, which limits the representation capacity. Another solution is to use dilated convolutions. However, this would lead to checkerboard artifacts \cite{dilate} in high-level feature maps. Our solution is simple, we use $16\times$ downsampling factor, but make two modifications. First, we add more layers with a smaller channel number in lower levels to extract more fine-detail information. Second, we adopt a top-down branch similar to FPN \cite {fpn} that combines high-resolution feature maps with low-resolution ones so as to up-sample the final feature representation. 

We show the backbone network architecture in Figure \ref{fig:network_architect}. To be specific, we have in total five blocks of layers in the backbone network. The first block consists of two convolutional layers with channel number 32 and stride 1. The second to fifth blocks are composed of residual layers \cite{resnet2} (with number of layers equals to 3, 6, 6, 3 respectively). The first convolution of each residual block has a stride of $2$ in order to down-sample the feature map. In total we have a down-sampling factor of $16$. To up-sample the feature map, we add a top-down path that up-samples the feature map by $2$ each time. This is then combined with the bottom-up feature maps at the corresponding resolution via pixel-wise summation. Two up-sampling layers are used, which leads to a final feature map with $4\times$ down-sampling factor with respect to the input.

\begin{figure}[t]
\begin{center}
   \includegraphics[width=0.5\linewidth]{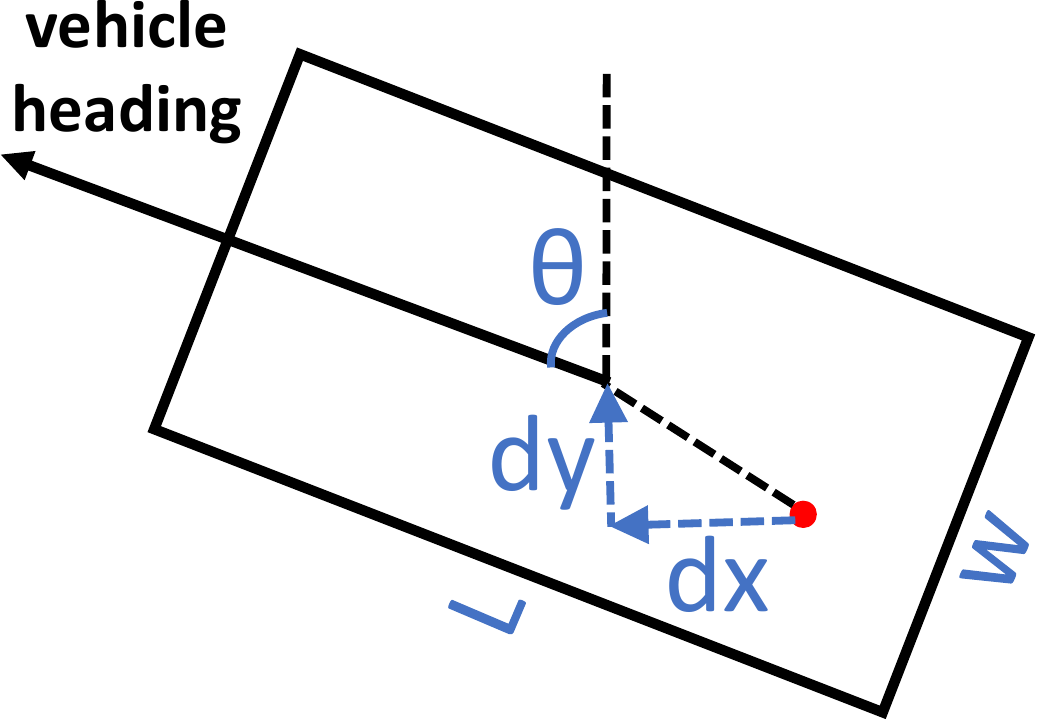}
\end{center}
   \caption{The geometry output parameterization for one positive sample (the red pixel). The learning target is $\{\cos(\theta), \sin(\theta), dx, dy, \log(w), \log(l)\}$, which is normalized before-hand over the training set to have zero mean and unit variance.}
\label{fig:output_representation}
\end{figure}

\begin{table*}[t!]
\begin{center}
\begin{small}
\addtolength{\tabcolsep}{-0pt}
\begin{tabular}{|c|r||ccc|c||c|c|c||c|c|c|}
\hline
\multirow{2}{*}{Method}& \multirow{2}{*}{Time (ms)}& \multicolumn{4}{|c||}{$AP_{0.7}$, val (\%)} & \multicolumn{3}{|c||}{$AP_{KITTI}$, val (\%)} & \multicolumn{3}{|c|}{$AP_{KITTI}$, test (\%)}\\
\cline{3-12}
& & 0-30 & 30-50 & 50-70 & 0-70 & Easy & \underline{Moderate} & Hard & Easy & \underline{Moderate} & Hard\\
\hline
VeloFCN~\cite{velofcn}  & 1000 & - & - & - & - & - & - & - & 0.15 & 0.33 & 0.47 \\
3D FCN~\cite{3dfcn}  & $>$5000 & - & - & - & - & - & - & - & 69.94 & 62.54 & 55.94 \\
MV3D~\cite{mv3d}  & 240 & 80.53 & 53.68 & 1.36 & 66.32 & 86.18 & 77.32 & 76.33 & 85.82 & 77.00 & 68.94 \\
MV3D+im~\cite{mv3d} & 360 & 76.16 & 58.41 & 4.87 & 65.31 & 86.55 & 78.10 & {\bf 76.67} & {\bf 86.02} & 76.90 & 68.49 \\
\hline
PIXOR & {\bf 35} & {\bf 87.68} & {\bf 60.05} & {\bf 21.62} & {\bf 75.74} & {\bf 86.79} & {\bf 80.75} & 76.60 & 81.70 & {\bf 77.05} & {\bf 72.95} \\
\hline
\end{tabular}
\end{small}
\caption{Evaluation results of LIDAR based 3D object detectors on KITTI BEV Object Detection validation and testing set. MV3D+im uses image as additional input. We use $AP_{0.7}$ (AUC of PR Curve with 0.7 IoU thresholds on all cars) and $AP_{KITTI}$ (official KITTI metric that computes the AUC with 11 sampling points only, evaluated on three sub-sets) as evaluation metrics. We also show fine-grained evaluation with regard to different ranges (distance in meters to the ego-car), which makes more sense in 3D detection.}
\label{tab:object_kitti}
\end{center}
\end{table*}

\subsubsection{Header Network}
The header network is a multi-task network that handles both object recognition and localization. It is designed to be small and efficient. The classification branch outputs 1-channel feature map followed with sigmoid activation function. The regression branch outputs 6-channel feature maps without non-linearity. There exists a trade-off in how many layers to share weights among the two branches. On the one hand, we'd like the weights to be utilized more efficiently. On the other hand, since they are different sub-tasks, we want them to be more separate and more specialized. We make an investigative experiment of this trade-off in next chapter, and find that sharing weights of the two tasks leads to slightly better performance.

We parameterize each object as an oriented bounding box $b$ as $\{\theta, x_c, y_c, w, l\}$, with each element corresponding to the heading angle (within range $[-\pi, \pi]$), the object's center position, and the object's size. Compared with cuboid based 3D object detection, we omit position and size along the $Z$ axis because in applications like autonomous driving the objects of interest are constrained to the same ground plane and therefore we only care about how to localize it on that plane (this setting is also known as 3D localization in some literatures \cite{mv3d}). Given such parameterization, the representation of the regression branch is $\{cos(\theta), sin(\theta), dx, dy, w, l\}$ for each pixel at position $(p_x, p_y)$ (shown as the red point in Figure \ref{fig:output_representation}). Note that the heading angle is factored into two correlated values to enforce the angle range constraint. We decode the $\theta$ as $atan2(sin(\theta), cos(\theta))$ during inference. $(dx, dy)$ corresponds to the position offset from the the pixel position to the object center. $(w, l)$ corresponds to the object size. It is worth noting that the values for the object position and size are in real-world metric space. The learning target is $\{\cos(\theta), \sin(\theta), dx, dy, \log(w), \log(l)\}$, which is normalized before-hand over the training set to have zero mean and unit variance.

\subsection{Learning and Inference}
We adopt the commonly used multi-task loss \cite{frcn} to train the full network. Specifically, we use cross-entropy loss on the classification output $p$ and a smooth $\ell_1$ loss on the regression output $q$. We sum the classification loss over all locations on the output map, while the regression loss is computed over positive locations only.
\begin{align*}
    & Loss = \textrm{focal\_loss}(p, y_{cls}) + \textrm{smooth}_{L_1}(q - y_{reg}) \\
    & \textrm{focal\_loss}(p, y) = 
    \begin{cases}
      -\alpha(1-p)^{\gamma}\text{log} (p)& \text{if } y = 1\\
      -(1-\alpha)p^{\gamma}\text{log} (1-p)& \text{otherwise},
    \end{cases} \\
    & \textrm{smooth}_{L_1}(x) =
    \begin{cases}
      0.5x^2& \text{if } |x| < 1\\
      |x| - 0.5& \text{otherwise},
    \end{cases}
\end{align*}
Note that we have severe class imbalance since a large proportion of the scene belongs to background. To stabilize the training process, we adopt the focal loss with the same hyper-parameter as \cite{focal} to re-weight all the samples. 
In the next chapter, we also propose a biased sampling strategy for positive samples that leads to better convergence. During inference, we feed the computed BEV representation from LIDAR point cloud to the network and get one channel of confidence score and six channels of geometry information. We then decode the geometry information into oriented bounding boxes only on positions whose confidence scores are above certain threshold. Non-Maximum-Suppression is used to get the final detections, where the overlap is computed as the Intersection-Over-Union of two oriented boxes.

%% file: results.tex

\begin{figure*}[t]
\begin{center}
   \includegraphics[width=1.0\linewidth]{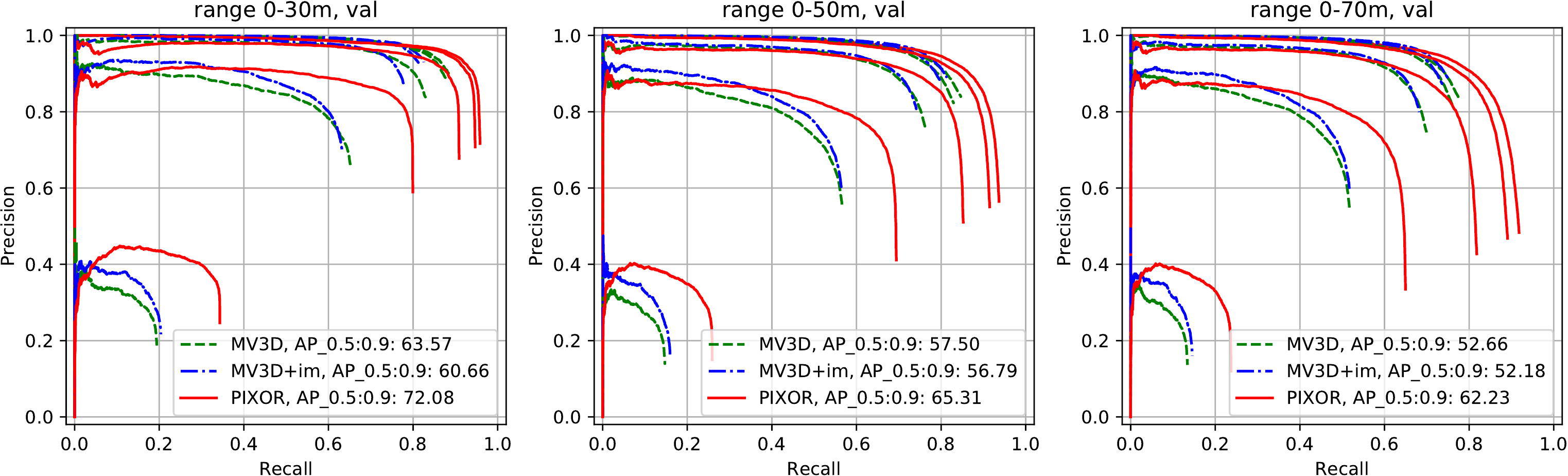}
\end{center}
\vspace*{-2mm}
\caption{Evaluation results of PIXOR and MV3D \cite{mv3d} on KITTI BEV Object Detection validation set. For each approach, we plot 5 Precision-Recall curves corresponding to 5 different IoU thresholds between 0.5 and 0.9, and report the averaged AP (\%). We compare in three different ranges.}
\label{fig:fine_kitti}
\vspace{-0.3cm}
\end{figure*}

\section{Experiments}

We conduct three types of experiments here. First, we compare PIXOR with other state-of-the-art 3D object detectors on the public KITTI bird's eye view object detection benchmark \cite{kitti}. We show that PIXOR achieves best performance both in accuracy and speed compared with all previously published methods. Second, we conduct an ablation study of PIXOR in three aspects: optimization, network architecture, and speed. Third, we verify the generalization ability of PIXOR by applying it to a new large-scale vehicle detection dataset for autonomous driving.

\subsection{BEV Object Detection on KITTI}

\subsubsection{Implementation Details} 
We set the region of interest for the point cloud to $[0, 70] \times [-40, 40]$ meters and do bird's eye view projection with a discretization resolution of $0.1$ meter. We set the height range to $[-2.5, 1]$ meters in LIDAR coordinates and divide all points into $35$ slices with bin size of $0.1$ meter. One reflectance channel is also computed. As a result, our input representation has the dimension of $800 \times 700 \times 38$. We use data augmentation of rotation between $[-5, 5]$ degrees along the Z axis and a random flip along $X$ axis during training. Unlike other detectors \cite{mv3d} that initialize the network weights from a pre-trained model, we train our network from scratch without resorting to any pre-trained model.

\vspace{-0.2cm}
\subsubsection{Evaluation Metric}
We use Average Precision (AP) computed at $0.7$ Intersection-Over-Union (IoU) as our evaluation metric in all experiments unless mentioned otherwise. We compute the AP as Area Under Precision-Recall Curve (AUC) \cite{pascal}. We evaluate on `Car' category and ignore `Van', `Truck', `Tram' and `DontCare' categories in KITTI during evaluation, meaning that we don't count True Positive (TP) or False Negative (FN) on them. Note that the metric we use is different from what KITTI reports in the following two aspects: (1) KITTI computes AP by sampling at $11$ linearly sampled recall rates (from $0\%$ to $100\%$), which is a rough approximation to real AUC. (2) KITTI divides labels into three subsets with image-based definition (e.g, object height in pixels, visibility in image), and reports AP on each subset, which doesn't suit pure LIDAR based object detection. In contrast, we evaluate on {\it all} labels within the region of interest, and do fine-grained evaluation with respect to ranges (object distance to the ego-car).

\vspace{-0.2cm}
\subsubsection{Evaluation Result}
We compare with 3D object detectors that use LIDAR on KITTI benchmark: VeloFCN \cite{velofcn}, 3D FCN \cite{3dfcn} and MV3D \cite{mv3d}. We show the evaluation results in Table \ref{tab:object_kitti}. From the table we see that PIXOR largely outperforms other approaches in AP at 0.7 IoU within 70 meters range, leading the second best by over $9\%$. We also show evaluation results with respect to ranges, and show that PIXOR outperforms more in the long range. On KITTI's test set, PIXOR outperforms MV3D in {\it moderate} and {\it hard} settings.

Since MV3D is the best approach among all state-of-the-art methods, we'd like to make a more detailed comparison using the AUC based AP metric. We show fine-grained Precision-Recall (PR) curves of both PIXOR and MV3D in Figure \ref{fig:fine_kitti}. From the figure, we get the following observations: (1) PIXOR outperforms MV3D in all IoU thresholds, especially at very high IoU like 0.8 and 0.9, showing that even without using proposal, PIXOR can still get super-accurate object localization, compared to the two-stage proposal based detector MV3D. (2) PIXOR has similar precision with MV3D at low recall rates. However, when it comes to higher recall rates, PIXOR shows huge advantage. At the same precision rate of the end point of MV3D's curve, PIXOR generally has over $5\%$ higher recall rate in all ranges. This shows that dense detector like PIXOR does have an advantage of higher recall rate, compared with two-stage detectors. (3) In the more difficult long range part, PIXOR still shows superiority over MV3D, which justifies our input representation design that reserves the 3D information well and our network architecture design that captures both fine details and regional context.

\begin{table}[t]
\begin{center}
\begin{small}
\begin{tabular}{|c|c|cc|}
\hline
Classification & Regression & $AP_{0.7}$ & $AP_{avg}$ \\
\hline
cross-entropy & smooth\_L1 & 73.46\% & 55.25\%\\
focal & smooth\_L1 & 74.93\% & 55.89\% \\
focal & decoding & 71.05\% & 53.05\% \\
focal & smooth\_L1 + decode (f.t.) & {\bf 77.16\%} & {\bf 58.31\%} \\
\hline
\end{tabular}
\caption{Ablation study of different loss functions. {\bf smooth\_L1 + decode (f.t.)} means that the network is trained with smooth L1 loss first, and then fine-tuned by replacing the smooth L1 loss with decoding loss.}
\label{tab:loss}
\end{small}
\end{center}
\vspace{-0.3cm}

\end{table}
\begin{table}[t]
\begin{center}
\begin{small}
\begin{tabular}{|c|c|cc|}
\hline
Training Samples & Data Aug. & $AP_{0.7}$ & $AP_{avg}$ \\
\hline
all pixels & none & 71.10\% & 53.99\% \\
ignore boundary pixels & none & 74.54\% & 55.79\% \\
ignore boundary pixels & rotate + flip & {\bf 74.93\%} & {\bf 55.89\%} \\
\hline
\end{tabular}
\caption{Ablation study of different data sampling strategies.}
\label{tab:sample}
\end{small}
\end{center}
\vspace{-0.5cm}
\end{table}

\subsection{Ablation Study}

We show an extensive ablation study of the proposed detector in terms of optimization, network architecture, speed and failure mode.

\vspace{-0.2cm}
\subsubsection{Experimental Setting}

Since we also compare with other state-of-the-art methods on the {\it val} set, it would be inappropriate to do the ablation study on the same set. Therefore we resort to KITTI Raw dataset \cite{kitti-raw} and randomly pick 3000 frames that are not overlapped with both {\it train} and {\it val} sets in KITTI object detection dataset, which we call {\it val-dev} set. We report ablation study results on this set. We use AP at 0.7 IoU as well as AP averaged from 0.5 to 0.95 IoUs (with a stride of 0.05) as the evaluation metrics.

\begin{table}[t]
\begin{center}
\begin{small}
\begin{tabular}{|l|c|}
\hline
Backbone Network & $AP_{avg}$ \\
\hline
pvanet & 51.28\% \\
resnet-50 & 53.03\% \\
vgg16-half & 54.46\% \\
ours & {\bf 55.07\%} \\
\hline
\end{tabular}
\caption{Ablation study of different backbone networks.}
\label{tab:backbone_exp}
\end{small}
\end{center}
\vspace{-0.5cm}
\end{table}
\begin{figure}[t]
\begin{center}
   \includegraphics[width=1.0\linewidth]{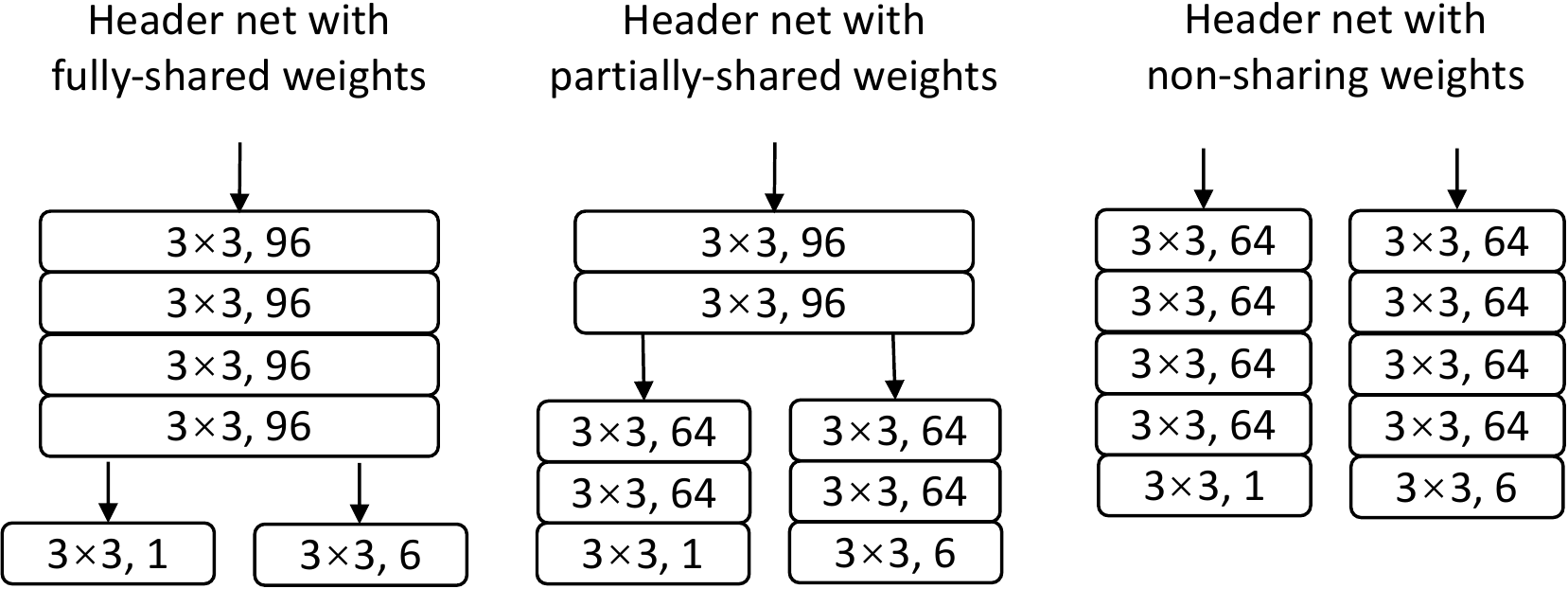}
\caption{Three versions of header network architectures.}
\label{fig:subnet}
\end{center}
\vspace{-0.5cm}
\end{figure}

\vspace{-0.2cm}
\subsubsection{Optimization}

We investigate into four topics here: the classification loss, the regression loss, the sampling strategy, and data augmentation.

\vspace{-0.2cm}
\paragraph{Classification loss} RetinaNet \cite{focal} proposes the focal loss to re-weight samples for dense detector training. For simplicity, we use their hyper-parameter setting. We show results in Table \ref{tab:loss}, and find that focal loss improves the AP\_0.7 by more than 1\%.

\vspace{-0.2cm}
\paragraph{Regression loss} For box regression, our default choice is smooth L1 loss \cite{frcn} on every dimension of the regression targets. We also adopt a {\it decoding loss}, where the output targets are first decoded into oriented boxes and then smooth L1 loss is computed on the (x, y) coordinates of four box corners directly with regard to ground-truth. Since the decoding the oriented box from regression targets is just a combination of some normal mathematic operations, this decoding process is differentiable and gradients can be back-propagated through this decoding process. We believe that this decoding loss is more end-to-end and implicitly balances different dimensions of the regression targets. In the results shown in Table \ref{tab:loss}, we show that directly training with the decoding loss doesn't work very well. However, training with conventional loss first and then fine-tuning with the proposed decoding loss helps improve the performance a lot.

\vspace{-0.2cm}
\paragraph{Data sampling and augmentation} 
When training dense detectors, one issue is how to define positive and negative samples. In proposal based approaches, this is defined by the IoU between proposal and ground-truth. Since PIXOR is a proposal-free method, we go for a more straightforward sampling strategy: all pixels inside the ground-truth are positive samples while outside pixels are negative samples. This simple definition already gives decent performance. However, one issue with this definition is that the variance of regression targets could be large for pixels near the object boundary. Therefore we propose to sub-sample the pixels, i.e, to ignore pixels near object boundary during training. Specifically, we zoom the ground-truth object box twice by $0.3\times$ and $1.2\times$ respectively, and ignore all pixels between these two. From the results shown in Table \ref{tab:sample}, we find that this sub-sampling strategy is beneficial to stabilize training. We also find that our data augmentation for KITTI helps a bit since PIXOR is trained from scratch instead of from a pre-trained model.

\begin{figure*}[t]
\begin{center}
   \includegraphics[width=0.24\linewidth]{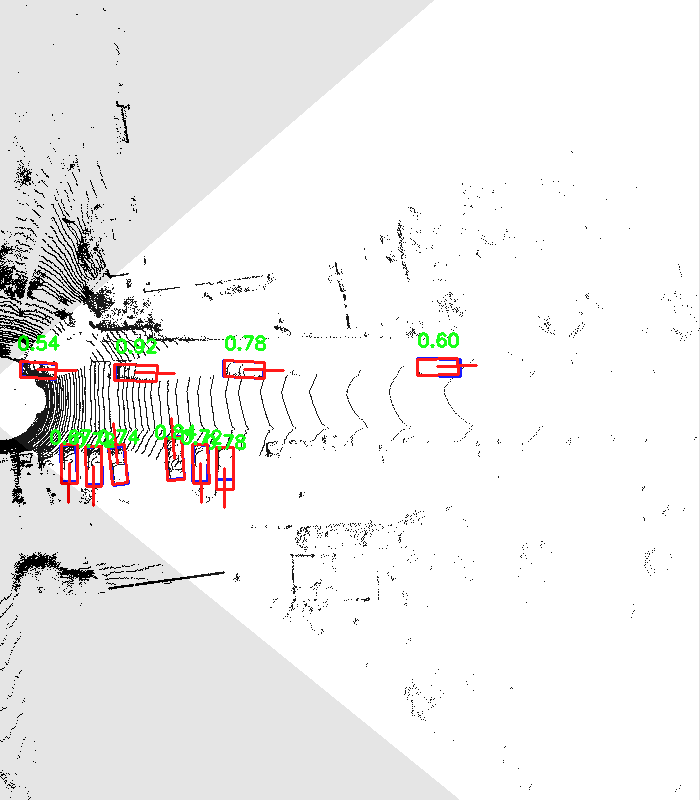}\includegraphics[width=0.24\linewidth]{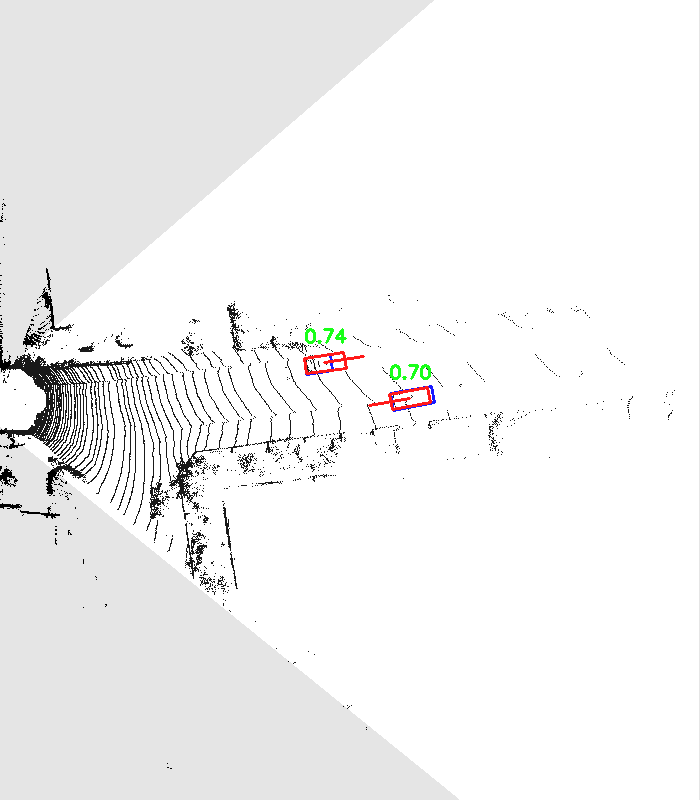}\includegraphics[width=0.24\linewidth]{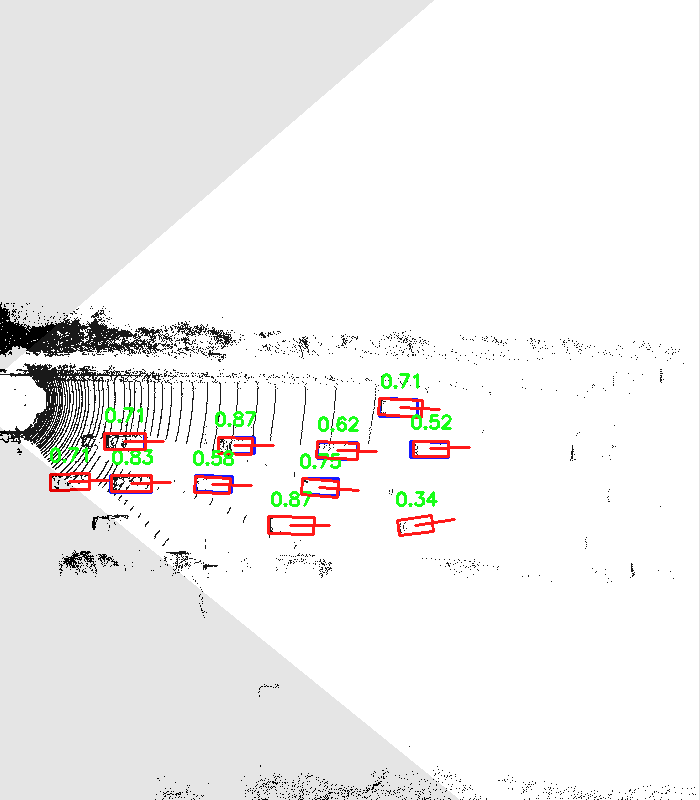}\includegraphics[width=0.24\linewidth]{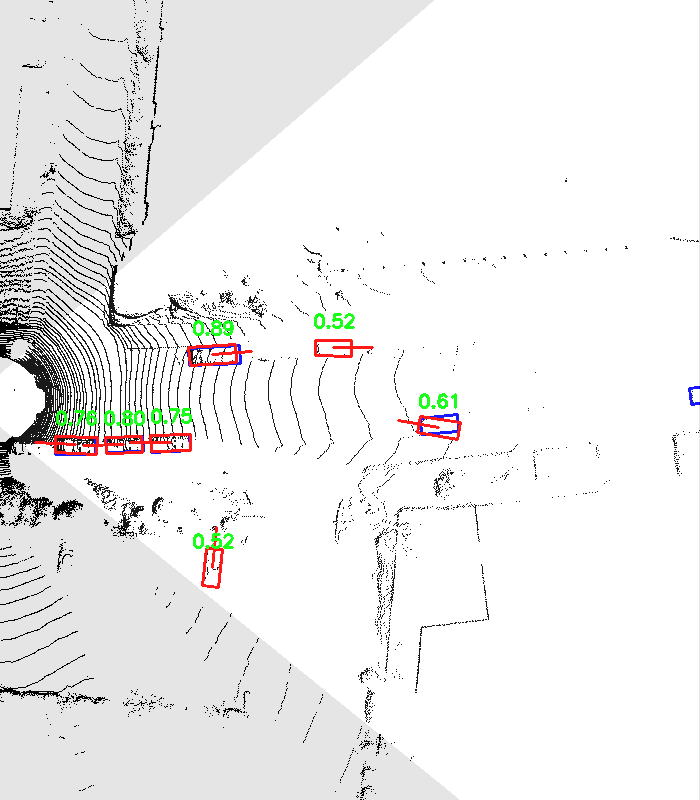}
   \includegraphics[width=0.24\linewidth]{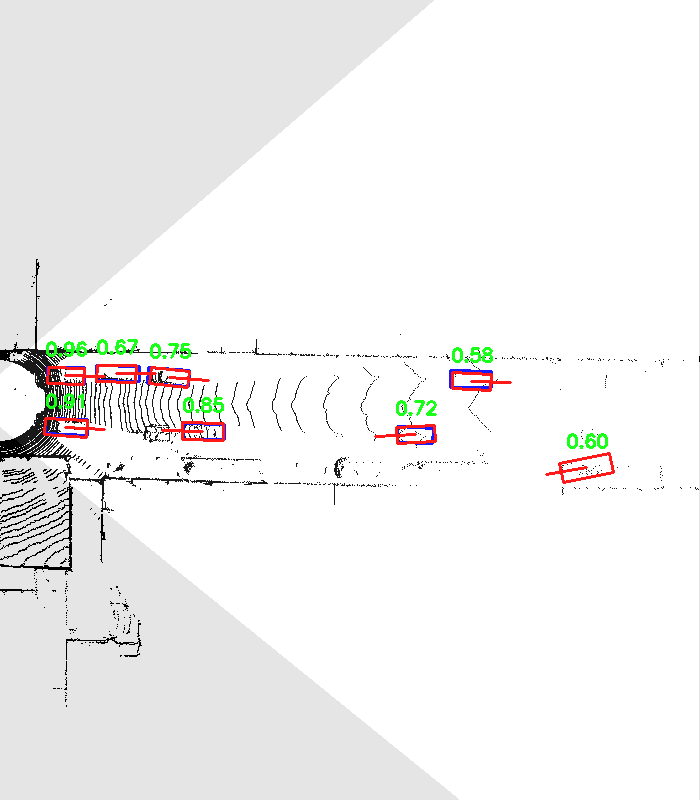}\includegraphics[width=0.24\linewidth]{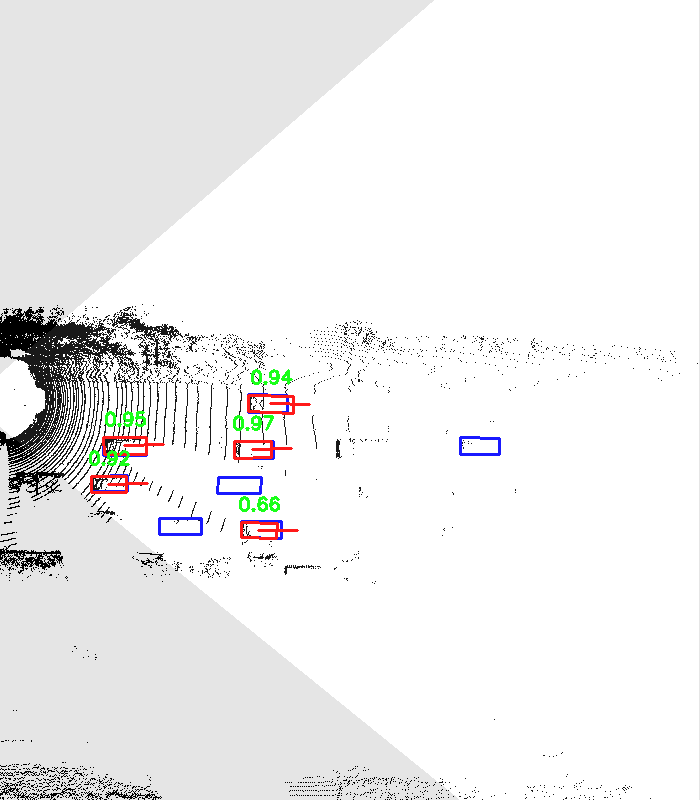}\includegraphics[width=0.24\linewidth]{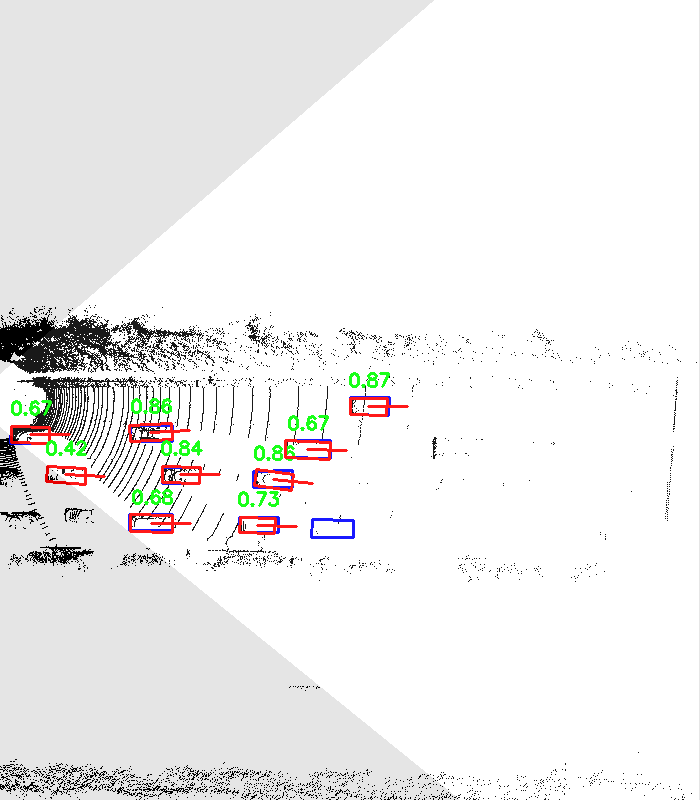}\includegraphics[width=0.24\linewidth]{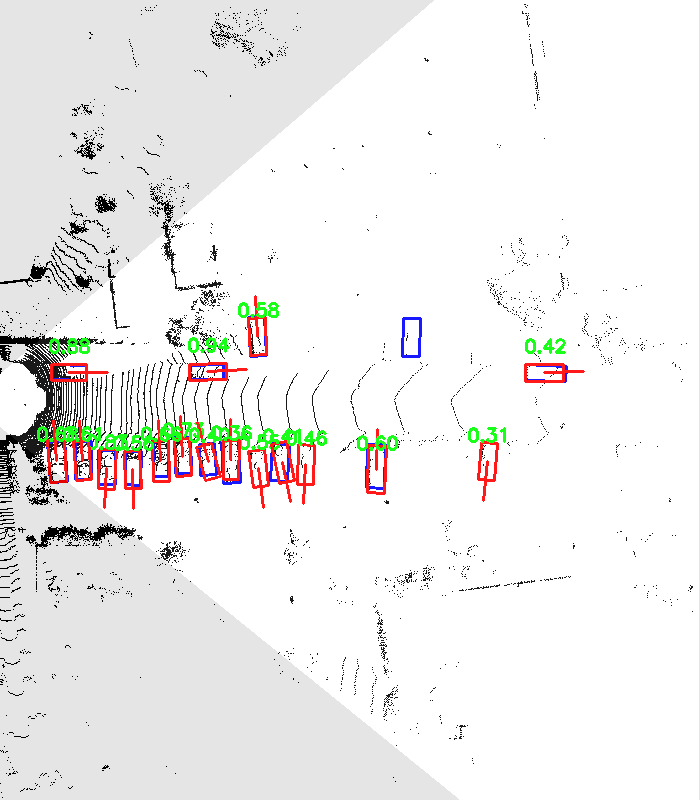}
\end{center}
\vspace*{-3mm}
   \caption{Example detection results of PIXOR on KITTI BEV Object Detection validation set. The detection is in red color, while the ground-truth is in blue color. Gray area is out of the scope of the camera view and therefore has no labels.}
\label{fig:det_demo}
\vspace{-0.3cm}
\end{figure*}
\begin{table}[t]
\begin{center}
\begin{small}
\begin{tabular}{|l|cc|}
\hline
Header Network & $AP_{0.7}$ & $AP_{avg}$ \\
\hline
non-sharing & 74.93\% & 55.89\% \\
partially-shared & 74.66\% & 55.75\% \\
fully-shared & {\bf 75.13\%} & {\bf 56.04\%} \\
\hline
\end{tabular}
\caption{Ablation study of different header network architectures.}
\label{tab:subnet}
\end{small}
\end{center}
\vspace{-0.5cm}
\end{table}
\begin{table}[t]
\begin{center}
\begin{small}
\begin{tabular}{|l|ccc|c|}
\hline
 & digitization & network & NMS & total \\
\hline
time (ms) & 1 & 31 & 3 & 35 \\
\hline
\end{tabular}
\caption{The detailed timing analysis of PIXOR on KITTI dataset.}
\label{tab:object_time}
\end{small}
\end{center}
\vspace{-0.5cm}
\end{table}


\vspace{-0.2cm}
\subsubsection{Network Architecture}

\vspace{-0.2cm}
\paragraph{Backbone network} We first compare different backbone networks: vgg16 with half channel number \cite{vgg}, pvanet \cite{pvanet}, resnet-50 \cite{resnet}, and the proposed architecture as presented in Figure \ref{fig:network_architect}. All of these backbone networks run below 100 milliseconds. All backbone networks except for {\it vgg16-half} uses residual unit as building blocks. We find that {\it vgg16-half} converges faster in {\it train} set and gets lower training loss than all other residual variants, but the performance drops quite a lot when evaluated on {\it val} set. This doesn't happen to the other three residual-based networks. We conjecture that this is because {\it vgg16-half} is more prone to overfitting without implicit regularization imposed by residual connections. 

\vspace{-0.2cm}
\paragraph{Header network} We also compare different structures for the header network. We investigate into how much we should share the parameters for the multi-task outputs. Three versions of header network are proposed with different extent of weight sharing in Figure \ref{fig:subnet} and compared in Table \ref{tab:subnet}. All these three versions have very close number of parameters. We find that fully-shared structure works best as it utilizes the parameters most efficiently.

\vspace{-0.2cm}
\subsubsection{Speed} We show detailed timing analysis of PIXOR in Table \ref{tab:object_time} for one single frame. All computations are performed on GPU. The network time is measured on a NVIDIA Titan Xp GPU and averaged over 100 non-sequential frames in KITTI.

\vspace{-0.2cm}
\subsubsection{Failure Mode} We show some detection results of PIXOR in Figure \ref{fig:det_demo}, and discover some failure modes. In general PIXOR will fail when there's no observed LIDAR points. In longer range we have very few evidence of the object, and therefore object localization becomes inaccurate, leading to false positives at higher IoU thresholds.

\subsection{BEV Object Detection on Large-scale Dataset}

\subsubsection{TOR4D Dataset}
We also collect a large-scale 3D vehicle detection dataset called TOR4D which has a different sensor configuration from KITTI and is collected in North-American cities. There are in total 6500 sequences collected, which are divided into 5000/500/1000 as train/val/test splits. The training sequences are sampled at 10 Hz into frames, while validation and testing sequences are sampled at 0.5Hz. As a result, there are over 1.2 million frames in training set, 5969 and 11969 frames in the val and test sets. All vehicles are annotated with bird's eye view bounding boxes.

\vspace{-0.2cm}
\subsubsection{Evaluation Result}
We make the following modifications to PIXOR on TOR4D dataset: we use ``vgg-half'' backbone network \cite{mv3d}, the detection region is 100m forward and backward and 40m to the left and right of the ego car, and the voxelization resolution is 0.2m. The network inference time of this model is 24 ms on a NVIDIA 1080Ti GPU. In comparison, we build a YOLO-like \cite{yolo} baseline detector with a customized backbone network, and add object anchors and multi-scale feature fusion to further improve the performance. The evaluation results on TOR4D test set are listed in Table \ref{tab:our-data}, where we show that PIXOR outperforms the baseline by 3.9\% in $AP_{0.7}$, proving that PIXOR is simple and easy to generalize.

\begin{table}[t]
\begin{center}
\begin{small}
\begin{tabular}{|l|c|}
\hline
Method & $AP_{0.7}$ \\
\hline
Baseline \cite{yolo} & 69.4\% \\
PIXOR & \textbf{73.3\%} \\
\hline
\end{tabular}
\caption{Evaluation of PIXOR on TOR4D test set.}
\label{tab:our-data}
\end{small}
\end{center}
\vspace{-0.7cm}
\end{table}

%% file: conc.tex

\section{Conclusion}
In this paper we propose a real-time 3D object detector called PIXOR that operates on LIDAR point clouds. PIXOR is a single-stage, proposal-free, dense object detector that achieve extreme simplicity in the context of 3D object localization for autonomous driving. PIXOR takes bird's eye view representation as input for efficiency in computation. We evaluate PIXOR on the challenging KITTI benchmark as well as a large-scale vehicle detection dataset TOR4D, and show that it outperforms the other methods by a large margin in terms of Average Precision (AP), while still runs at $>28$ FPS.

%% file: ack.tex

\section*{Acknowledgement}

We thank Gregory P. Meyer for suggesting the decoding loss, Andrei Pokrovsky for GPU implementation of oriented NMS, and the anonymous reviewers for their insightful suggestions.

%% file: top.bbl
\begin{thebibliography}{10}\itemsep=-1pt

\bibitem{mono3d}
X.~Chen, K.~Kundu, Z.~Zhang, H.~Ma, S.~Fidler, and R.~Urtasun.
\newblock Monocular 3d object detection for autonomous driving.
\newblock In {\em Proceedings of the IEEE Conference on Computer Vision and
  Pattern Recognition}, pages 2147--2156, 2016.

\bibitem{3doppami}
X.~Chen, K.~Kundu, Y.~Zhu, H.~Ma, S.~Fidler, and R.~Urtasun.
\newblock 3d object proposals using stereo imagery for accurate object class
  detection.
\newblock {\em IEEE Transactions on Pattern Analysis and Machine Intelligence},
  2017.

\bibitem{mv3d}
X.~Chen, H.~Ma, J.~Wan, B.~Li, and T.~Xia.
\newblock Multi-view 3d object detection network for autonomous driving.
\newblock In {\em Proceedings of the IEEE Conference on Computer Vision and
  Pattern Recognition}, 2017.

\bibitem{rfcn}
J.~Dai, Y.~Li, K.~He, and J.~Sun.
\newblock R-fcn: Object detection via region-based fully convolutional
  networks.
\newblock In {\em Advances in Neural Information Processing Systems}, pages
  379--387, 2016.

\bibitem{deng2009imagenet}
J.~Deng, W.~Dong, R.~Socher, L.-J. Li, K.~Li, and L.~Fei-Fei.
\newblock Imagenet: A large-scale hierarchical image database.
\newblock In {\em Computer Vision and Pattern Recognition, 2009. CVPR 2009.
  IEEE Conference on}, pages 248--255. IEEE, 2009.

\bibitem{vote3deep}
M.~Engelcke, D.~Rao, D.~Z. Wang, C.~H. Tong, and I.~Posner.
\newblock Vote3deep: Fast object detection in 3d point clouds using efficient
  convolutional neural networks.
\newblock In {\em Robotics and Automation (ICRA), 2017 IEEE International
  Conference on}, pages 1355--1361. IEEE, 2017.

\bibitem{everingham2010pascal}
M.~Everingham, L.~Van~Gool, C.~K. Williams, J.~Winn, and A.~Zisserman.
\newblock The pascal visual object classes (voc) challenge.
\newblock {\em International journal of computer vision}, 88(2):303--338, 2010.

\bibitem{pascal}
M.~Everingham, L.~Van~Gool, C.~K. Williams, J.~Winn, and A.~Zisserman.
\newblock The pascal visual object classes (voc) challenge.
\newblock {\em International Journal of Computer Vision}, 88(2):303--338, 2010.

\bibitem{kitti-raw}
A.~Geiger, P.~Lenz, C.~Stiller, and R.~Urtasun.
\newblock Vision meets robotics: The kitti dataset.
\newblock {\em The International Journal of Robotics Research},
  32(11):1231--1237, 2013.

\bibitem{kitti}
A.~Geiger, P.~Lenz, and R.~Urtasun.
\newblock Are we ready for autonomous driving? the kitti vision benchmark
  suite.
\newblock In {\em Computer Vision and Pattern Recognition (CVPR), 2012 IEEE
  Conference on}, pages 3354--3361. IEEE, 2012.

\bibitem{frcn}
R.~Girshick.
\newblock Fast r-cnn.
\newblock In {\em Proceedings of the IEEE International Conference on Computer
  Vision}, pages 1440--1448, 2015.

\bibitem{rcnn}
R.~Girshick, J.~Donahue, T.~Darrell, and J.~Malik.
\newblock Rich feature hierarchies for accurate object detection and semantic
  segmentation.
\newblock In {\em Proceedings of the IEEE Conference on Computer Vision and
  Pattern Recognition}, pages 580--587, 2014.

\bibitem{spp}
K.~He, X.~Zhang, S.~Ren, and J.~Sun.
\newblock Spatial pyramid pooling in deep convolutional networks for visual
  recognition.
\newblock {\em IEEE transactions on pattern analysis and machine intelligence},
  37(9):1904--1916, 2015.

\bibitem{resnet}
K.~He, X.~Zhang, S.~Ren, and J.~Sun.
\newblock Deep residual learning for image recognition.
\newblock In {\em Proceedings of the IEEE Conference on Computer Vision and
  Pattern Recognition}, pages 770--778, 2016.

\bibitem{resnet2}
K.~He, X.~Zhang, S.~Ren, and J.~Sun.
\newblock Identity mappings in deep residual networks.
\newblock In {\em European Conference on Computer Vision}, pages 630--645.
  Springer, 2016.

\bibitem{pvanet}
S.~Hong, B.~Roh, K.-H. Kim, Y.~Cheon, and M.~Park.
\newblock {PVANet}: Lightweight deep neural networks for real-time object
  detection.
\newblock {\em arXiv preprint arXiv:1611.08588}, 2016.

\bibitem{densebox}
L.~Huang, Y.~Yang, Y.~Deng, and Y.~Yu.
\newblock Densebox: Unifying landmark localization with end to end object
  detection.
\newblock {\em arXiv preprint arXiv:1509.04874}, 2015.

\bibitem{alexnet}
A.~Krizhevsky, I.~Sutskever, and G.~E. Hinton.
\newblock Imagenet classification with deep convolutional neural networks.
\newblock In {\em Advances in neural information processing systems}, pages
  1097--1105, 2012.

\bibitem{3dfcn}
B.~Li.
\newblock 3d fully convolutional network for vehicle detection in point cloud.
\newblock In {\em Intelligent Robots and Systems (IROS), 2017 IEEE/RSJ
  International Conference on}, pages 1513--1518. IEEE, 2017.

\bibitem{velofcn}
B.~Li, T.~Zhang, and T.~Xia.
\newblock Vehicle detection from 3d lidar using fully convolutional network.
\newblock In {\em Robotics: Science and Systems}, 2016.

\bibitem{fpn}
T.-Y. Lin, P.~Doll{\'a}r, R.~Girshick, K.~He, B.~Hariharan, and S.~Belongie.
\newblock Feature pyramid networks for object detection.
\newblock In {\em Proceedings of the IEEE Conference on Computer Vision and
  Pattern Recognition}, 2017.

\bibitem{focal}
T.-Y. Lin, P.~Goyal, R.~Girshick, K.~He, and P.~Doll{\'a}r.
\newblock Focal loss for dense object detection.
\newblock In {\em Proceedings of the IEEE International Conference on Computer
  Vision}, 2017.

\bibitem{ssd}
W.~Liu, D.~Anguelov, D.~Erhan, C.~Szegedy, S.~Reed, C.-Y. Fu, and A.~C. Berg.
\newblock Ssd: Single shot multibox detector.
\newblock In {\em European Conference on Computer Vision}, pages 21--37.
  Springer, 2016.

\bibitem{long2015fully}
J.~Long, E.~Shelhamer, and T.~Darrell.
\newblock Fully convolutional networks for semantic segmentation.
\newblock In {\em Proceedings of the IEEE Conference on Computer Vision and
  Pattern Recognition}, pages 3431--3440, 2015.

\bibitem{dilate}
A.~Odena, V.~Dumoulin, and C.~Olah.
\newblock Deconvolution and checkerboard artifacts.
\newblock {\em Distill}, 1(10):e3, 2016.

\bibitem{mcg}
J.~Pont-Tuset, P.~Arbelaez, J.~T. Barron, F.~Marques, and J.~Malik.
\newblock Multiscale combinatorial grouping for image segmentation and object
  proposal generation.
\newblock {\em IEEE transactions on Pattern Analysis and Machine Intelligence},
  39(1):128--140, 2017.

\bibitem{yolo}
J.~Redmon, S.~Divvala, R.~Girshick, and A.~Farhadi.
\newblock You only look once: Unified, real-time object detection.
\newblock In {\em Proceedings of the IEEE Conference on Computer Vision and
  Pattern Recognition}, pages 779--788, 2016.

\bibitem{faster}
S.~Ren, K.~He, R.~Girshick, and J.~Sun.
\newblock Faster r-cnn: Towards real-time object detection with region proposal
  networks.
\newblock In {\em Advances in Neural Information Processing Systems}, pages
  91--99, 2015.

\bibitem{russakovsky2015imagenet}
O.~Russakovsky, J.~Deng, H.~Su, J.~Krause, S.~Satheesh, S.~Ma, Z.~Huang,
  A.~Karpathy, A.~Khosla, M.~Bernstein, et~al.
\newblock Imagenet large scale visual recognition challenge.
\newblock {\em International Journal of Computer Vision}, 115(3):211--252,
  2015.

\bibitem{overfeat}
P.~Sermanet, D.~Eigen, X.~Zhang, M.~Mathieu, R.~Fergus, and Y.~LeCun.
\newblock Overfeat: Integrated recognition, localization and detection using
  convolutional networks.
\newblock {\em International Conference on Learning Representations (ICLR
  2014)}, 2014.

\bibitem{vgg}
K.~Simonyan and A.~Zisserman.
\newblock Very deep convolutional networks for large-scale image recognition.
\newblock {\em CoRR}, abs/1409.1556, 2014.

\bibitem{ss}
S.~Song and J.~Xiao.
\newblock Sliding shapes for 3d object detection in depth images.
\newblock In {\em European conference on computer vision}, pages 634--651.
  Springer, 2014.

\bibitem{dss}
S.~Song and J.~Xiao.
\newblock Deep sliding shapes for amodal 3d object detection in rgb-d images.
\newblock In {\em Proceedings of the IEEE Conference on Computer Vision and
  Pattern Recognition}, pages 808--816, 2016.

\bibitem{suykens1999least}
J.~A. Suykens and J.~Vandewalle.
\newblock Least squares support vector machine classifiers.
\newblock {\em Neural processing letters}, 9(3):293--300, 1999.

\bibitem{tran2015learning}
D.~Tran, L.~Bourdev, R.~Fergus, L.~Torresani, and M.~Paluri.
\newblock Learning spatiotemporal features with 3d convolutional networks.
\newblock In {\em Proceedings of the IEEE international conference on computer
  vision}, pages 4489--4497, 2015.

\bibitem{selective}
J.~R. Uijlings, K.~E. Van De~Sande, T.~Gevers, and A.~W. Smeulders.
\newblock Selective search for object recognition.
\newblock {\em International Journal of Computer Vision}, 104(2):154--171,
  2013.

\bibitem{vote3d}
D.~Z. Wang and I.~Posner.
\newblock Voting for voting in online point cloud object detection.
\newblock In {\em Robotics: Science and Systems}, 2015.

\bibitem{east}
X.~Zhou, C.~Yao, H.~Wen, Y.~Wang, S.~Zhou, W.~He, and J.~Liang.
\newblock East: An efficient and accurate scene text detector.
\newblock In {\em Proceedings of the IEEE Conference on Computer Vision and
  Pattern Recognition}, 2017.

\end{thebibliography}
